%% file: emnlp16aae.tex
\documentclass[11pt,letterpaper]{article}
\usepackage{emnlp2016}
\usepackage[round]{natbib}   
\usepackage{times}
\usepackage{latexsym}
\usepackage{amsmath}
\usepackage{graphicx}

\usepackage[usenames,dvipsnames]{color}
\definecolor{myblue}{rgb}{0,0.1,0.6}
\definecolor{mygreen}{rgb}{0,0.3,0.1}
\usepackage[colorlinks=true,linkcolor=black,citecolor=mygreen,urlcolor=myblue]{hyperref}

\newcommand{\bocomment}[1]{}
\newcommand{\scomment}[1]{}

\newenvironment{itemizesquish}{\begin{list}{\labelitemi}{\setlength{\itemsep}{0em}\setlength{\labelwidth}{2em}\setlength{\leftmargin}{\labelwidth}\addtolength{\leftmargin}{\labelsep}}}{\end{list}}
\newcommand{\ignore}[1]{}

\emnlpfinalcopy

\newcommand{\fullversion}[1]{#1}

\def\emnlppaperid{1039}

\title{
Demographic Dialectal Variation in Social Media:
A Case Study of African-American English}

\author{Su Lin Blodgett$^\dag$
    \ \  Lisa Green$^*$
    \ \ Brendan O'Connor$^\dag$ \\
    $^\dag$College of Information and Computer Sciences
    \hspace{1.5em}
    $^*$Department of Linguistics
    \\
    University of Massachusetts Amherst
  }

\date{}

\begin{document}

\maketitle

\begin{abstract}
Though dialectal language is increasingly abundant on social media, 
few resources exist for developing NLP tools to handle such language.
We conduct a case study of dialectal language in online conversational text by investigating African-American English (AAE) on Twitter. We propose a distantly supervised model to identify AAE-like language from demographics associated with geo-located messages, and we verify that this language follows well-known AAE linguistic phenomena. In addition, we analyze the quality of existing language identification and dependency parsing tools on AAE-like text, demonstrating that they perform poorly on such text compared to text associated with white speakers. We also provide an ensemble classifier for language identification which eliminates this disparity and release a new corpus of tweets containing AAE-like language.  

Data and software resources are available at: \\
\url{http://slanglab.cs.umass.edu/TwitterAAE}
\fullversion{

\noindent \emph{(This is an expanded version of our EMNLP 2016 paper, including the appendix at end.)}}
\end{abstract}

\section{Introduction}


Owing to variation within a standard language, regional and social dialects exist within languages across the world. These varieties or dialects differ from the standard variety in syntax (sentence structure), phonology (sound structure), and the inventory of words and phrases (lexicon). Dialect communities often align with geographic and sociological factors, as language variation emerges within distinct social networks, or is affirmed as a marker of social identity. 

As many of these dialects have traditionally existed primarily in oral contexts, they have historically been underrepresented in written sources. Consequently, NLP tools have been developed from text which aligns with mainstream languages. With the rise of social media, however, dialectal language is playing an increasingly prominent role in online conversational text, for which traditional NLP tools may be insufficient.  This impacts many applications: for example, dialect speakers' opinions may be mischaracterized under social media sentiment analysis or omitted altogether \citep{hovy2016social}.
Since this data is now available, we seek to analyze current NLP challenges
and extract dialectal language from online data.

Specifically, we investigate dialectal language in publicly available Twitter data, focusing on African-American English (AAE), a dialect of Standard American English (SAE) spoken by millions of people across the United States. AAE is a linguistic variety with defined syntactic-semantic, phonological, and lexical features, which have been the subject of a rich body of sociolinguistic literature. In addition to the linguistic characterization, reference to its speakers and their geographical location or speech communities is important, especially in light of the historical development of the dialect. Not all African-Americans speak AAE, and not all speakers of AAE are African-American; nevertheless, speakers of this variety have close ties with specific communities of African-Americans \citep{Green2002AAE}. Due to its widespread use, established history in the sociolinguistic literature, and demographic associations, AAE provides an ideal starting point for the development of a statistical model that uncovers dialectal language.
In fact, its presence in social media is attracting increasing interest
for natural language processing \citep{Jorgensen2016AAE}
and sociolinguistic \citep{stewart2014now,Eisenstein2015Review,jones2015toward}
research.\footnote{Including a recent linguistics workshop: \url{http://linguistlaura.blogspot.co.uk/2016/06/using-twitter-for-linguistic-research.html}} In this work we:

\begin{itemizesquish}
\item Develop a method to identify \emph{demographically-aligned} text and language from geo-located messages (\S\ref{s:method}),
based on distant supervision of geographic census demographics 
through a statistical model that assumes a soft correlation between demographics and language.
\item Validate our approach by 
verifying that
text aligned with African-American demographics
follows well-known phonological and syntactic properties of AAE,
and document the previously unattested ways in which such text diverges from SAE (\S\ref{s:ling}).
\item Demonstrate \emph{racial disparity} in the efficacy of NLP tools 
for language identification and dependency parsing---they perform poorly on this text, compared to text associated with white speakers (\S\ref{s:langid}, \S\ref{s:parse}).
\item Improve language identification for U.S.\ online conversational text with a simple ensemble classifier using our demographically-based distant supervision method, aiming to eliminate racial disparity in accuracy rates (\S\ref{s:langidfix}).
\item Provide a corpus of 830,000 tweets aligned with African-American demographics.
\end{itemizesquish}

\noindent

\section{Identifying AAE from Demographics} \label{s:method}

\noindent The presence of AAE in social media and the generation of resources of AAE-like text for NLP tasks has attracted recent interest in sociolinguistic and natural language processing research; \cite{jones2015toward} shows that nonstandard AAE orthography on Twitter aligns with 
historical patterns of African-American migration in the U.S., while \cite{jorgensen2015challenges} investigate to what extent it supports well-known sociolinguistics hypotheses about AAE. Both, however, find AAE-like language on Twitter through keyword searches, which may not yield broad corpora reflective of general AAE use. 
More recently, \cite{Jorgensen2016AAE} generated a large unlabeled corpus of text from hip-hop lyrics, subtitles from \emph{The Wire} and \emph{The Boondocks}, and tweets from a region of the southeast U.S.
While this corpus does indeed capture a wide variety of language, we aim to discover AAE-like language 
by utilizing finer-grained, neighborhood-level demographics from across the country.

Our approach to identifying AAE-like text is to first
harvest a set of messages from Twitter, cross-referenced against U.S.\ Census demographics (\S\ref{s:twcensus}), then to analyze words against demographics with two alternative methods, a seedlist approach (\S\ref{s:seedlist}) and a mixed-membership probabilistic model (\S\ref{s:model}).

\subsection{Twitter and Census data} \label{s:twcensus}

\noindent In order to create a corpus of demographically-associated dialectal language, we turn to Twitter, whose public messages contain large amounts of casual conversation and dialectal speech \citep{Eisenstein2015Review}.
It is well-established that Twitter can be used to study
both geographic dialectal varieties\footnote{For example, of American English \citep{huang2015understanding,doyle2014mapping}.}
and minority languages.\footnote{For example, \cite{lynn2015minority} develop POS corpora and taggers for Irish tweets; see also related work in \S\ref{s:langideval}.}

Some methods exist to associate messages with authors' races;
one possibility is to use birth record statistics to identify African-American-associated
names, which has been used in (non-social media)
social science studies \citep{sweeney2013discrimination,bertrand2003emily}.
However, metadata about authors is fairly limited on Twitter and most other social media services, and many supplied names are obviously not real.

Instead, we turn to geo-location and induce a distantly supervised mapping between authors and the demographics of the neighborhoods they live in \citep{OConnor2010Demo,Eisenstein2011SparseSoc,stewart2014now}.
We draw on a set of geo-located Twitter messages, most of which are sent on mobile phones, by authors in the U.S. in 2013. (These are selected from a general archive of the
``Gardenhose/Decahose'' sample stream of public Twitter messages \citep{morstatter2013sample}).
Geo-located users are a particular sample of the userbase \citep{pavalanathan2015confounds}, but we expect
it is reasonable to compare users of different races within this group.

We look up the U.S.\ Census blockgroup geographic area that the message was sent in; blockgroups are one of the smallest geographic areas defined by the Census, 
typically containing a population of 600--3000 people.
We use race and ethnicity information for each blockgroup from the Census' 2013 American Community Survey, 
defining four covariates: percentages of the population
that are non-Hispanic whites, non-Hispanic blacks, Hispanics (of any race), and Asian.\footnote{See appendix for additional details.}  Finally, for each user $u$, we average the demographic values of all their messages in our dataset into a length-four vector $\pi^{(census)}_u$.
Under strong assumptions, this could be interpreted as the probability of which race the user is; we prefer to think of it as a rough proxy for likely demographics of the author and the neighborhood they live in.

Messages were filtered in order to focus on casual conversational text; 
we exclude tweets whose authors had 1000 or more followers, or that
(a) contained 3 or more hashtags,
(b) contained the strings ``http'', ``follow'', or ``mention'' (messages designed to generate followers), or (c) were retweeted (either containing the string ``rt'' or marked by Twitter's metadata as re-tweeted).

Our initial Gardenhose/Decahose stream archive had 16 billion messages in 2013; 90 million were geo-located with coordinates that matched a U.S.\ Census blockgroup. 59.2 million tweets from 2.8 million users remained after pre-processing; each user is associated with a set of messages and averaged demographics $\pi^{(census)}_u$.



\subsection{Direct Word-Demographic Analysis} \label{s:seedlist}

\noindent Given a set of messages and demographics associated with their authors, a number of methods could be used to infer statistical associations between language and demographics.

Direct word-demographic analysis methods use the $\pi^{(census)}_u$ quantities 
to 
calculate statistics at the word level in a single pass.
An intuitive approach is to calculate the \emph{average demographics per word}.
For a token in the corpus indexed by $t$ (across the whole corpus),
let $u(t)$ be the author of the message containing that token, and $w_t$ be the word token.  The average demographics of word type $w$ is:\footnote{
$\pi_{w,k}$ has the flavor of ``soft counts'' in multinomial EM. By changing the
denominator to $\sum_t \pi^{(census)}_{u(t)}$,
it calculates a unigram
language model that sums to one across the vocabulary.
This hints at a more complete modeling approach (\S\ref{s:model}).
}
\[ \pi^{(softcount)}_w \equiv \frac{\sum_t 1\{w_t=w\} \pi^{(census)}_{u(t)}} {\sum_t 1\{w_t=w\}}\]
We find that terms with the highest $\pi_{w,\text{AA}}$ values (denoting high average African-American demographics of their authors' locations) are very non-standard, while
\cite{stewart2014now} and \cite{eisenstein2013phonological} find large $\pi_{w,\text{AA}}$ associated with certain AAE linguistic features.

One way to use the $\pi_{w,k}$ values to construct a corpus is through a seedlist approach.
In early experiments, we constructed a corpus of 41,774 users (2.3 million messages)
by first
selecting the $n=100$ highest-$\pi_{w,\text{AA}}$ terms occurring at least $m=3000$ times across the data set, then collecting all tweets from frequent authors who have at least 10 tweets
and frequently use these terms,
defined as the case when at least $p=20\%$ of their messages contain
at least one of the seedlist terms.  Unfortunately, the $n,m,p$ thresholds are ad-hoc.

\ignore{
This would corresond to the first step of an EM algorithm for a language model for each demographic category, where posterior demographic memberships were initialized as $\pi^{(census)}$ priors.  Further EM steps would require specification of a generative process in order to re-calculate posterior demographic memberships, which would allow inference in cases where,
for example, an author in a low-AA place, but who uses AA-associated language, could be inferred as an author who does in fact use AA-associated language.  To properly consider joint evidence in this manner, we turn to a fully specified probabilistic model in the next section.
}


\subsection{Mixed-Membership Demographic-Language Model} \label{s:model}

\noindent The direct word-demographics analysis gives useful validation that the demographic information may yield dialectal corpora, and the seedlist approach can assemble a set of users 
with heavy dialectal usage.  However, the approach requires a number of ad-hoc thresholds, cannot capture authors who only occasionally use demographically-aligned language, and cannot differentiate
language use at the message-level.  To address these concerns, we develop a mixed-membership model
for demographics and language use in social media.

\begin{figure}
\centering
\includegraphics[width=2in]{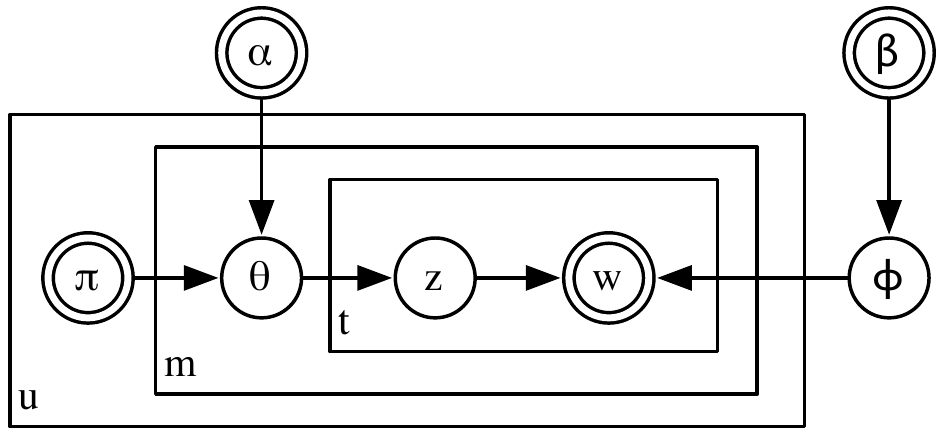}
\begin{align*}
    \theta_m &\sim Dir(\alpha \pi_u),\ \ 
    \phi \sim Dir(\beta/V) \\
    z_t &\sim \theta_m,\ \ w_z \sim \phi_{z_t}
\end{align*}
\vspace{-0.3in}
\caption{Mixed-membership model for users ($u$), messages ($m$) and tokens ($t$). 
Observed variables have a double lined border. \label{f:plate}}
\end{figure}

The model directly associates each of the four demographic variables with a topic; i.e.\ a unigram language model over the vocabulary.\footnote{To build the vocabulary,
we select all words used by at least 20 different users, resulting in 191,873 unique words; other words are mapped to an out-of-vocabulary symbol.}
The model assumes an author's mixture over the topics tends to be similar to their Census-associated demographic weights, and that every message has its own topic distribution.
This allows for a single author to use different types of language in different messages,
accommodating multidialectal authors.
The message-level topic probabilities $\theta_m$
are drawn from an asymmetric Dirichlet centered on $\pi^{(census)}_u$,
whose scalar concentration parameter $\alpha$ controls whether authors' language is very similar to the demographic prior, or can have some deviation.  A token $t$'s latent topic $z_t$ 
is drawn from $\theta_m$, and the word itself is drawn from $\phi_{z_t}$, the language model for the topic (Figure~\ref{f:plate}).

Thus the model learns \emph{demographically-aligned} language models for each demographic category.
The model is much more tightly constrained than a topic model---for example, if $\alpha \rightarrow \infty$, $\theta$ becomes fixed and the likelihood is concave as a function of $\phi$---but it still has more joint learning than a direct calculation approach,
since the inference of a messages' topic memberships $\theta_m$ 
is affected not just by the Census priors, but also by the language used.  A tweet written by an author in a highly AA neighborhood may be inferred to be non-AAE-aligned if it uses non-AAE-associated terms; as inference proceeeds, this information is used to learn sharper language models.

We fit the model with collapsed Gibbs sampling \citep{Griffiths2004CGS} with repeated sample updates for each token $t$ in the corpus,
\[ p(z_t=k \mid w,z_{-t}) \propto
\frac{N_{wk}+\beta/V}{N_k+\beta}
\frac{N_{mk}+\alpha \pi_{uk}}{N_m+\alpha}
\]
where $N_{wk}$ is the number of tokens where word $w$ occurs under topic $z=k$, $N_{mk}$ is the number of tokens in the current message with topic $k$, etc.; all counts exclude the current $t$ position.
We observed convergence of the log-likelihood within 100 to 200 iterations, and ran for 300 total.\footnote{Our straightforward single core implementation (in Julia)
spends 80 seconds for each iteration over 586 million tokens.}
We average together count tables from the last 50 Gibbs samples for
analysis of posterior topic memberships at the word, message, and user level;
for example, the posterior probability a particular user $u$ uses topic $k$, $P(z=k \mid u)$, can be calculated as the fraction of tokens with topic $k$ within messages authored by $u$.

We considered $\alpha$ to be a fixed control parameter; setting it higher increases the correlations between $P(z=k \mid u)$ and $\pi^{(census)}_{u,k}$.
We view the selection of $\alpha$ as an inherently difficult problem,
since the correlation between race and AAE usage is already complicated and imperfect
at the author-level, and census demographics allow only for rough associations.
We set $\alpha=10$ which yields posterior user-level correlations of 
$P(z=AA \mid u)$ against $\pi_{u,AA}$ to be approximately $0.8$.

This model has broadly similar goals as non-latent, log-linear generative models of text that condition on document-level covariates \citep{Monroe2008Fightin,Eisenstein2011SAGE,Taddy2013MNIR}.
The formulation here has the advantage of fast inference with large vocabularies (since the partition function never has to be computed), and gives probabilistic admixture semantics
at arbitrary levels of the data.
This model is also related to topic models where 
the selection of $\theta$ conditions on covariates
\citep{Mimno2008DMR,Ramage2011Partial,Roberts2013STMNIPS},
though it is much simpler without full latent topic learning.

In early experiments, we used only two classes (AA and not AA), and found Spanish terms being included in the AA topic.  Thus we turned to four race categories in order to better draw out non-AAE language.  This removed Spanish terms from the AA topic; interestingly, they did not go to the Hispanic topic, but instead to Asian, along with other foreign languages.  
In fact, the correlation between users' Census-derived proportions of Asian populations,
versus this posterior topic's proportions,
is only 0.29, while the other three topics correlate to their respective Census priors in the range 0.83 to 0.87.
This indicates the ``Asian'' topic actually functions as a background topic (at least in part).
Better modeling of demographics and non-English language interactions
is interesting potential future work.

By fitting the model to data, we can directly analyze unigram probabilities within the model parameters $\phi$,
but for other analyses, such as analyzing larger syntactic constructions and testing NLP tools, we require an explicit corpus of messages. 

To generate a user-based AA-aligned corpus, we collected all tweets from users whose posterior probability of using AA-associated terms under the model was at least 80\%,
and generated a corresponding white-aligned corpus as well.
In order to remove the effects of non-English languages, and given uncertainty about 
what the model learned in the Hispanic and Asian-aligned demographic topics,
we focused only on AA- and white-aligned language
by imposing the additional constraint that each user's combined posterior 
proportion of Hispanic or Asian language was less than 5\%.
Our two resulting user corpora contain 830,000 and 7.3 million tweets,
for which we are making their message IDs available for further research
(in conformance with the Twitter API's Terms of Service).
In the rest of the work, we refer to these as the AA- and white-aligned corpora, respectively.

\section{Linguistic Validation} \label{s:ling}

\noindent Because validation by manual inspection of our AA-aligned text is impractical, we turn to the well-studied phonological and syntactic phenomena that traditionally distinguish AAE from SAE. We validate our model by reproducing these phenomena, and document a variety of other ways in which our AA-aligned text diverges from SAE.

\subsection{Lexical-Level Variation} \label{s:lex}




\noindent We begin by examining how much AA- and white-aligned lexical items diverge from a standard dictionary. We used SCOWL's largest wordlist with level 1 variants as our dictionary, totaling 627,685 words.\footnote{http://wordlist.aspell.net/}

We calculated, for each word $w$ in the model's vocabulary, the ratio

\begin{equation*}
    r_{k}(w) = \frac{p(w|z=k)}{p(w|z \neq k)}
\end{equation*}
where the $p(.|.)$ probabilities are posterior inferences, derived from averaged Gibbs samples
of the sufficient statistic count tables $N_{wk}$.

We selected heavily AA- and white-aligned words as those where $r_{AA}(w) \geq 2$ and $r_{white}(w) \geq 2$, respectively. We find that while 58.2\% of heavily white-aligned words were not in our dictionary, fully 79.1\% of heavily AA-aligned words were not. While a high number of out-of-dictionary lexical items is expected for Twitter data, this disparity suggests that the AA-aligned lexicon diverges from SAE
more strongly than the white-aligned lexicon. 


\subsection{Internet-Specific Orthographic Variation} \label{s:inet}


\noindent We performed an ``open vocabulary" unigram analysis by ranking all words in the vocabulary by $r_{AA}(w)$ and browsed them and samples of their usage.
Among the words with high $r_{AA}$, we observe a number of Internet-specific orthographic variations, which we separate into three types: abbreviations (e.g. \emph{llh}, \emph{kmsl}), shortenings (e.g. \emph{dwn}, \emph{dnt}), and spelling variations which do not correlate to the word's pronunciation (e.g. \emph{axx}, \emph{bxtch}). These variations do not reflect features attested in the literature; rather, they appear to be purely orthographic variations highly specific to AAE-speaking communities online. They may highlight previously unknown linguistic phenomena; for example, we observe that \emph{thoe} (SAE \emph{though}) frequently appears in the role of a discourse marker instead of its standard SAE usage (e.g. \emph{Girl Madison outfit THOE}). This new use of \emph{though} as a discourse marker, which is difficult to observe using the SAE spelling amidst many instances of the SAE usage, is readily identifiable in examples containing the \emph{thoe} variant. Thus, non-standard spellings provide valuable windows into a variety of linguistic phenomena.

In the next section, we turn to variations which do appear to arise from known phonological processes.


\subsection{Phonological Variation} \label{s:phon}

\noindent Many phonological features are closely associated with AAE \citep{Green2002AAE}. While there is not a perfect correlation between orthographic variations and people's pronunciations, \cite{eisenstein2013phonological} shows that some genuine phonological phenomena, including a number of AAE features, are accurately reflected in orthographic variation on social media.
We therefore validate our model by verifying that spellings reflecting known AAE phonological features align closely with the AA topic.


We selected 31 variants of SAE words from previous studies of AAE phonology on Twitter \citep{jorgensen2015challenges,jones2015toward}. These variations display a range of attested AAE phonological features, such as derhotacization (e.g. \emph{brotha}), deletion of initial \emph{g} and \emph{d} (e.g. \emph{iont}), and realization of voiced \emph{th} as \emph{d} (e.g. \emph{dey}) \citep{rickford1999african}.

Table \ref{t:ratios} shows the top five of these words by their $r_{AA}(w)$ ratio.
For 30 of the 31 words, $r \geq 1$, and for 13 words, $r \geq 100$, suggesting that our model strongly identifies words displaying AAE phonological features with the AA topic. The sole exception is the word \emph{brotha}, which appears to have been adopted into general usage as its own lexical item.

\begin{table}
\begin{center}
\begin{tabular}{|c|c|c|} \hline
\textbf{AAE} & \textbf{Ratio}  & \textbf{SAE} \\ \hline
sholl & 1802.49 & sure  \\ \hline
iont  & 930.98 & I don't \\ \hline
wea   & 870.45 & where  \\ \hline
talmbout & 809.79 & talking about \\ \hline
sumn   & 520.96 & something \\ \hline
\end{tabular}
\end{center}
\caption{Of 31 phonological variant words, top five by ratio $r_{AA}(w)$. SAE translations are shown for reference.}
\label{t:ratios}
\end{table}

\subsection{Syntactic Variation} \label{s:syntax}

\noindent We further validate our model by verifying that it reproduces well-known AAE syntactic constructions,
investigating three well-attested AAE aspectual or preverbal markers: 
habitual \emph{be}, 
future \emph{gone}, 
and completive \emph{done} 
\citep{Green2002AAE}.  Table \ref{t:patterns} shows examples of each construction.

To search for the constructions, we tagged the corpora using
the ARK Twitter POS tagger \citep{Gimpel2011POS,Owoputi2013POS},\footnote{Version 0.3.2: \url{http://www.cs.cmu.edu/~ark/TweetNLP/}}
which \cite{jorgensen2015challenges} show has similar accuracy rates on both AAE and non-AAE tweets,
unlike other POS taggers. We searched for each construction by searching for sequences of unigrams and POS tags characterizing the construction; e.g. for habitual \emph{be} we searched for the sequences O-\emph{be}-V and O-\emph{b}-V. Non-standard spellings for the unigrams in the patterns were identified from the ranked analysis of \S\ref{s:inet}.


\begin{table}
\begin{center}
\begin{tabular}{|p{2.6cm}|p{3cm}|c|} \hline
\textbf{Construction} & \textbf{Example} & \textbf{Ratio} \\ \hline
O-\emph{be/b}-V & \emph{I be tripping bruh} & 11.94 \\ \hline
\emph{gone/gne/gon}-V & \emph{Then she gon be single Af} & 14.26\\ \hline
\emph{done/dne}-V & \emph{I done laughed so hard that I'm weak} & 8.68\\ \hline
\end{tabular}
\caption{AAE syntactic constructions and the ratios of their occurrences in the AA- vs. white-aligned corpora (\S\ref{s:model}). \vspace{-0.3in}}
\label{t:patterns}
\end{center}
\end{table}




We examined how a message's likelihood of using each construction varies with the message's posterior probability of AA. We split all messages into deciles based on the messages' posterior probability of AA. From each decile, we sampled 200,000 messages and calculated the proportion of messages containing the three syntactic constructions.

For all three constructions, we observed the clear pattern that as messages' posterior probabilities of AA increase, so does their likelihood of containing the construction. Interestingly, for all three constructions, frequency of usage peaks at approximately the [0.7, 0.8) decile.
One possible reason for the decline in higher deciles
might be tendency of high-AA messages to be shorter; while the mean number of tokens per message across all deciles in our samples is 9.4, the means for the last two deciles are 8.6 and 7.1, respectively.

\begin{figure}
    \centering\vspace{-0.15in}
    \includegraphics[width=3.15in]{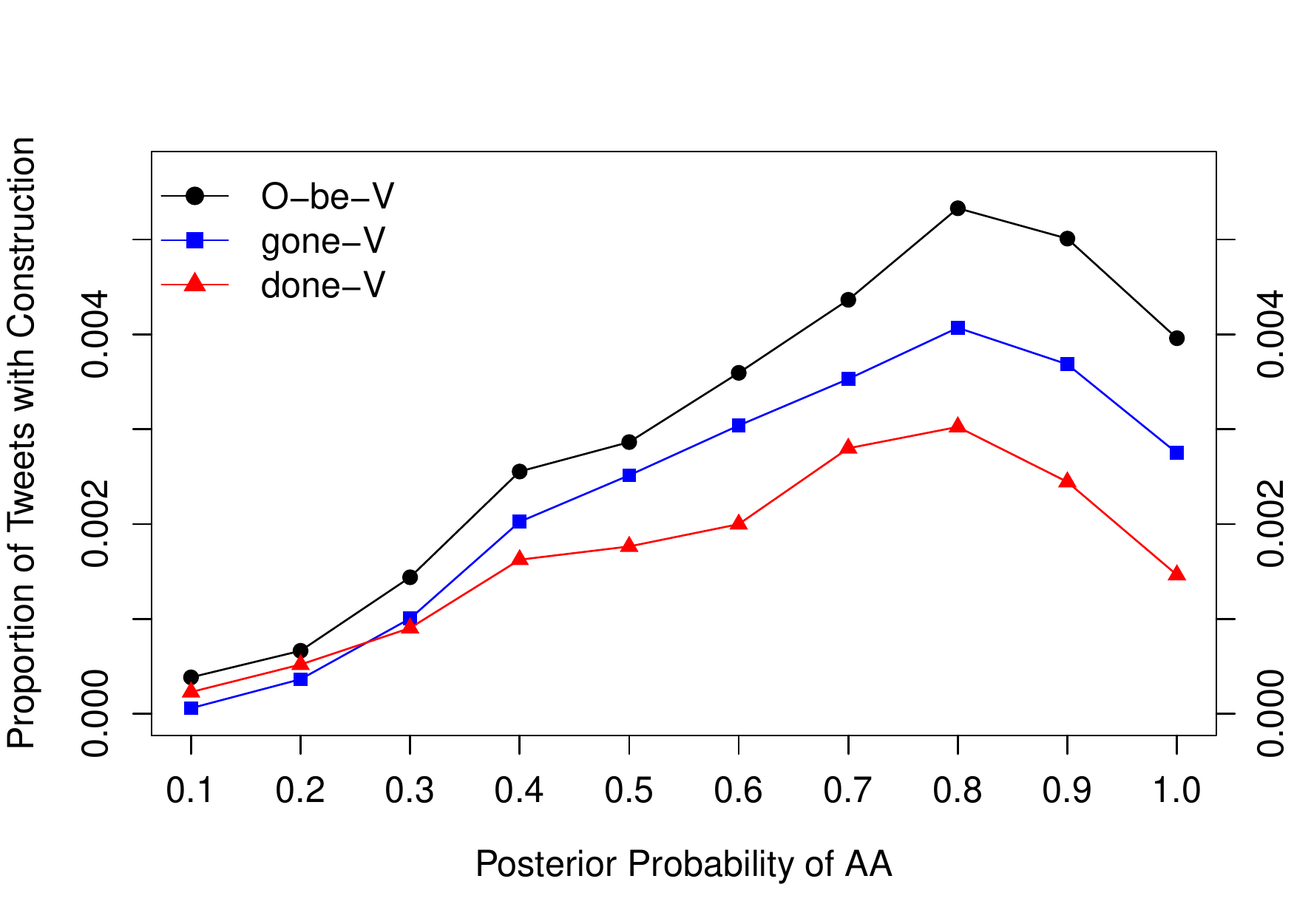}
    \caption{Proportion of tweets containing AAE syntactic constructions by messages' posterior probability of AA. On the x-axis, 0.1 refers to the decile [0, 0.1). }
    \label{f:syntax}
\end{figure}



Given the important linguistic differences between our demographically-aligned subcorpora, we hypothesize that current NLP tools may behave differently.
We investigate this hypothesis in \S\ref{s:langid} and \S\ref{s:parse}.

\section{Lang ID Tools on AAE} \label{s:langid}

\subsection{Evaluation of Existing Classifiers} \label{s:langideval}



\begin{table}
\begin{center}
\begin{tabular}{c|cc} \hline
& \textbf{AAE} & \textbf{White-Aligned} \\ \hline
\textbf{\emph{langid.py}} & 13.2\% & 7.6\% \\
\textbf{Twitter-1} & 8.4\% & 5.9\% \\
\textbf{Twitter-2} & 24.4\% & 17.6\% \\ \hline
\end{tabular}
\end{center}
\caption{Proportion of tweets in AA- and white-aligned corpora classified as non-English by different classifiers. (\S\ref{s:langideval})
}
\label{t:pcts}
\end{table}

Language identification, the task of classifying the major world language in which a message is written, is a crucial first step in almost any web or social media text processing pipeline.  For example, in order to analyze the opinions of U.S.\ Twitter users, one might throw away all non-English messages before running an English sentiment analyzer. 

\cite{hughes2006reconsidering} review 
language identification methods;
social media 
language identification is challenging
since messages are short, and also use non-standard and multiple (often related) languages \citep{baldwin2013noisy}.
Researchers have sought to model code-switching in
social media language \citep{rosner2007tagging,solorio2008learning,maharjan2015developing,zampieri2013n,king2013labeling},
and recent workshops have focused on code-switching \citep{solorio-EtAl:2014:CodeSwitch} 
and general language identification \citep{zubiaga-EtAl:2014:tweet}.
For Arabic dialect classification, work has developed corpora in both traditional and Romanized script \citep{cotterell2014algerian,malmasi2015arabic}
 and tools that use n-gram and morphological analysis to identify code-switching between dialects and with English \citep{elfardy2014aida}. 


We take the perspective that since AAE is a dialect of American English,
it ought to be classified as English for the task of major world language identification.
\cite{Lui2012Langid} develop
 \emph{langid.py}, one of the most popular open source language identification tools,
training it
on over 97 languages from texts including Wikipedia, and evaluating
on both traditional corpora and Twitter messages.
We hypothesize that if a language identification tool
is trained on standard English data,
it may exhibit disparate performance on AA- versus white-aligned tweets.  Since language identifiers are typically based on character n-gram features,
they may get confused by the types of
lexical/orthographic divergences
seen in \S\ref{s:ling}.
To evaluate this hypothesis, we compare the behavior of existing language identifiers on our subcorpora.
 
We test \emph{langid.py} as well as the output of Twitter's in-house identifier,
whose predictions are included in a tweet's metadata 
(from 2013, the time of data collection);
the latter may give a language code or a missing value (\emph{unk} or an empty/null value).
We record the proportion of non-English predictions by these systems; \emph{Twitter-1} does not consider missing values to be a non-English prediction, and \emph{Twitter-2} does.

We noticed emojis had seemingly unintended consequences on \emph{langid.py}'s classifications,
so removed all emojis by characters
from the relevant Unicode ranges.
We also removed @-mentions.

\textbf{User-level analysis} We begin by comparing the classifiers' behavior on the AA- and white-aligned corpora. Of the AA-aligned tweets, 13.2\% were classified by \textit{langid.py} as non-English; in contrast, 7.6\% of white-aligned tweets were classified as such. We observed similar disparities for \emph{Twitter-1} and \emph{Twitter-2},
illustrated in
Table~\ref{t:pcts}.

It turns out these ``non-English'' tweets
are, for the most part, actually English.
We sampled and annotated 50 tweets from the tweets classified as non-English by each run. Of these 300 tweets, only 3 could be unambiguously identified as written in a language other than English.

\textbf{Message-level analysis} We examine how a message's likelihood of being classified as non-English varies with its posterior probability of AA. As in \S\ref{s:syntax}, we split all messages into deciles based on the messages' posterior probability of AA, and predicted language identifications
on 200,000 sampled messages from each decile.

For all three systems, the proportion of messages classified as non-English increases steadily as the messages' posterior probabilities of AA increase. As before, we sampled and annotated from the tweets classified as non-English, sampling 50 tweets from each decile for each of the three systems. Of the 1500 sampled tweets, only 13 ($\sim$0.87\%) could be unambiguously identified as being in a language other than English.

\begin{figure}
\begin{center}
\vspace{-0.3in}
\includegraphics[width=3.15in]{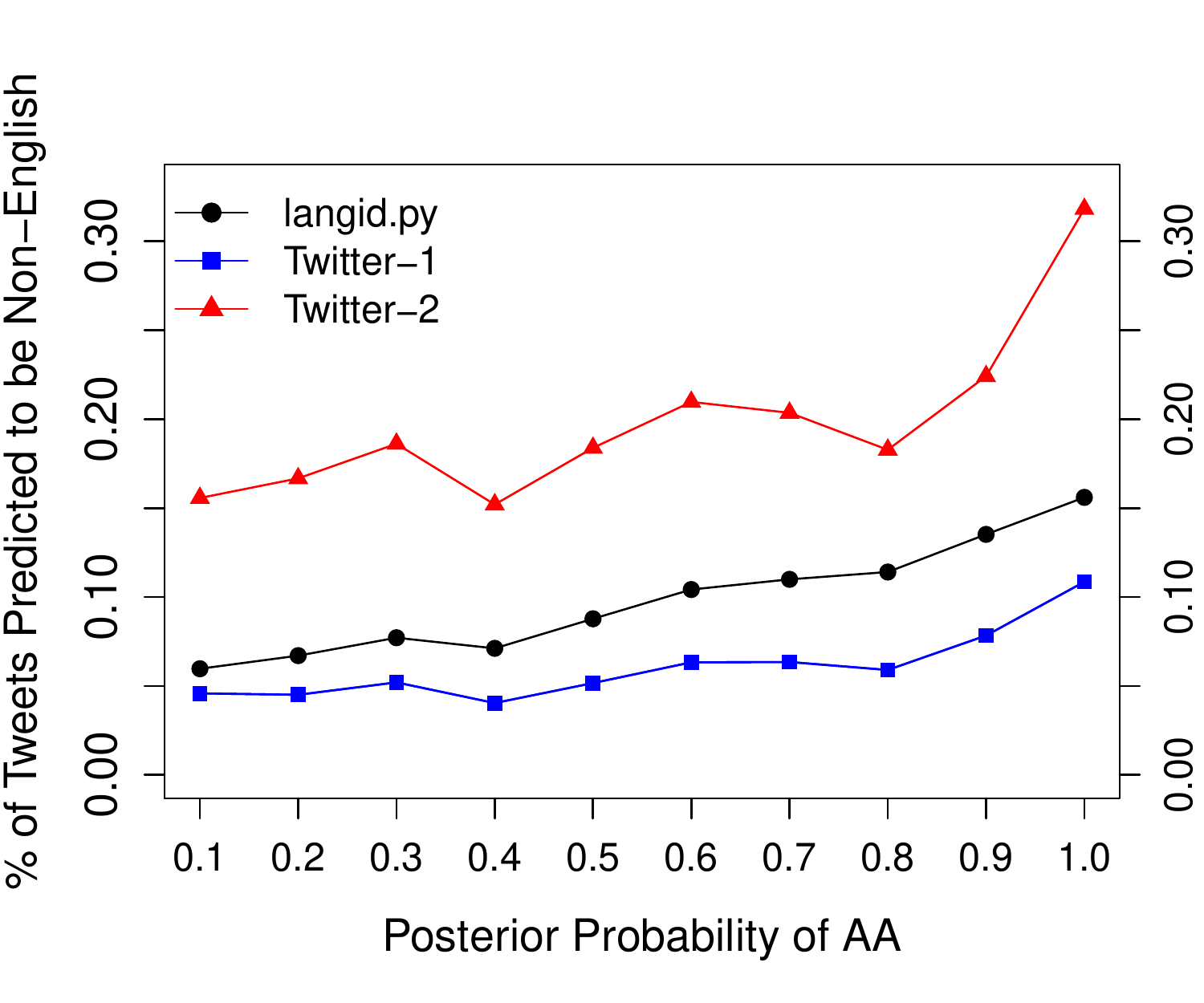}
\end{center}
\vspace{-0.25in}
\caption{Proportion of tweets classified as non-English by messages' posterior probability of AA. On the x-axis, 0.1 refers to the decile [0, 0.1).}
\label{f:langid}
\end{figure}

\subsection{Adapting Language Identification for AAE} \label{s:langidfix}

\noindent Natural language processing tools can be improved to better support dialects;
for example,
\cite{Jorgensen2016AAE} use domain adaptation methods
to improve POS tagging on AAE corpora.
In this section, we contribute a fix to language identification to correctly identify AAE and other social media messages as English.

\subsubsection{Ensemble Classifier}

\noindent 
We observed that messages where our model infers
a high probability of AAE, white-aligned, or ``Hispanic''-aligned language almost always are written in English; therefore we construct a simple ensemble classifier by combining it with \emph{langid.py}.

For a new message $\vec{w}$, we predict its demographic-language proportions $\hat{\theta}$ via posterior inference with our trained model, given a symmetric $\alpha$ prior over demographic-topic proportions (see appendix for details).
The ensemble classifier, given a message, is as follows:
\vspace{-0.03in}
\begin{itemizesquish}
\item Calculate langid.py's prediction $\hat{y}$.
\item If $\hat{y}$ is English, accept it as English.
\item If $\hat{y}$ is non-English, and at least one of the message's tokens are in demographic model's vocabulary:
Infer $\hat{\theta}$ and return English only if the combined AA, Hispanic, and white posterior probabilities are at least 0.9. Otherwise return the non-English $\hat{y}$ decision.
\end{itemizesquish}
Another way to view this method is that we are effectively training a system on an extended Twitter-specific English language corpus softly labeled by our system's posterior inference; in this respect, it is related to efforts to collect new
language-specific Twitter corpora \citep{bergsma2012language}
or minority language data from the web
\citep{ghani2001mining}.

\vspace{-0.02in}
\subsubsection{Evaluation}

\begin{table}
\begin{center}
\begin{tabular}{|c|c|c|} \hline
\textbf{Message set} & \textbf{\emph{langid.py}} & \textbf{Ensemble} \\ \hline
High AA & 80.1\% & 99.5\% \\
High White & 96.8\% & 99.9\% \\ 
\emph{General} & \emph{88.0\%} & \emph{93.4\%} \\ \hline
\end{tabular}
\end{center}
\caption{Imputed recall of English messages in 2014 messages.
For the \emph{General} set these are an approximation; see text. \vspace{-0.2in}}
\label{t:recall}
\end{table}

\noindent Our analysis from \S\ref{s:langideval} indicates that this method would
correct erroneous false negatives for AAE messages in the training set for the model.
We further confirm this
by testing the classifier on 
a sample of 2.2 million geolocated tweets sent in the U.S. in 2014,
which are not in the training set.

In addition to performance on the entire sample, we examine our classifier's performance on messages whose posterior probability of using AA- or white-associated terms was greater than 0.8 within the sample, which in this section we will call high AA and high white messages, respectively. Our classifier's precision is high across the board, at 100\% 
across manually annotated samples of 200 messages from each sample.\footnote{We annotated 600 messages
as English, not English, or not applicable, from 200 sampled each from 
general, high AA, and high white messages. 
Ambiguous tweets which were too short (e.g.\ "Gm") or contained only named entities (e.g. "Tennessee") were excluded from the final calculations.
The resulting samples have 197/197, 198/198, and 200/200 correct English classifications, respectively.}
Since we are concerned about the system's overall recall, we impute recall (Table~\ref{t:recall})
by assuming that all high AA and high white messages are indeed English.
Recall for \emph{langid.py} alone is calculated by $\frac{n}{N}$, where $n$ is the number of messages predicted to be English by \emph{langid.py}, and $N$ is the total number of messages in the set. 
(This is the complement of Table~\ref{t:pcts}, except evaluated on the test set.)
We estimate the ensemble's recall as $\frac{n + m}{N}$, where 
$m = (n_{flip}) P(\textrm{English} \mid flip)$ is the expected number of correctly changed classifications
(from non-English to English) by the ensemble and the second term is the precision (estimated as 1.0).
We observe the baseline system has considerable difference in recall between the groups
which is solved by the ensemble.

We also apply the same calculation to the general set of all 2.2 million messages;
the baseline classifies 88\% as English.  This is a less accurate approximation of recall
since we have observed
a substantial presence of non-English messages.
The ensemble classifies an additional 5.4\% 
of the messages as English; since these are all (or nearly all) correct, this reflects at least a 5.4\% gain to recall.

\section{Dependency Parser Evaluation} \label{s:parse}

\noindent Given the lexical and syntactic variation of AAE compared to SAE,
we hypothesize that syntactic analysis tools 
also have differential accuracy.
\cite{jorgensen2015challenges} demonstrate this
for part-of-speech
tagging, finding that SAE-trained taggers had disparate accuracy on AAE versus non-AAE tweets.

We assess a publicly available syntactic dependency parser
on our AAE and white-aligned corpora.
Syntactic parsing for tweets has received some research attention;
\cite{Foster2011Parsing} create a corpus of constituent trees for English tweets,
and
\cite{Kong2014Tweeboparser}'s \emph{Tweeboparser}
is trained on a Twitter corpus annotated with a customized unlabeled dependency formalism;
since its data was uniformly sampled from tweets, 
we expect it may have low disparity between demographic groups.

We focus on widely used syntactic representations,
testing the \emph{SyntaxNet}
neural network-based
dependency parser \citep{Andor2016Parseface},\footnote{Using the publicly available  
\emph{mcparseface} model: \url{https://github.com/tensorflow/models/tree/master/syntaxnet}
}
which reports state-of-the-art results, including for web corpora.
We evaluate it
against a new manual annotation of 200 messages, 100 randomly sampled from each of the AA- and white-aligned corpora described in \S\ref{s:model}.

\emph{SyntaxNet} outputs grammatical relations conforming to
the Stanford Dependencies (SD) system \citep{Marneffe2008DepManual},
which we used to annotate messages using \emph{Brat},\footnote{\url{http://brat.nlplab.org/}}
comparing to predicted parses for reference.
Message order was randomized and demographic inferences
were hidden from the annotator.
To increase statistical power relative to annotation effort,
we developed a partial annotation approach
to only
annotate edges for the root word of the first major sentence
in a message.
Generally, we found that that SD worked well as a descriptive formalism for tweets' syntax;
we describe handling of AAE and Internet-specific non-standard issues
in the appendix. 
\newcommand{\se}[1]{{\footnotesize (#1)}}
We evaluate labeled recall of the annotated edges
for each message set:
\vspace{-0.03in}
\begin{center}
    \begin{tabular}{rccc}
  Parser  & AA & Wh. & Difference \\
    \hline
   SyntaxNet  & 64.0 \se{2.5} & 80.4 \se{2.2} & 16.3 \se{3.4} \\
   CoreNLP  & 50.0 \se{2.7} &  71.0 \se{2.5} &  21.0 \se{3.7} \\
    \hline
    \end{tabular}
\end{center}
Bootstrapped standard errors (from 10,000 message resamplings) are in parentheses;
differences are statistically significant
($p<10^{-6}$ in both cases).

The white-aligned accuracy rate of 80.4\% is broadly in line with previous work
(compare to the parser's unlabeled accuracy of 89\% on English Web Treebank full annotations),
but parse quality is much worse on AAE tweets at 64.0\%.
We test the Stanford CoreNLP neural network dependency parser 
\citep{Chen2014Dep}
using the \emph{english\_SD} model that outputs this formalism;\footnote{\emph{pos,depparse} options in version 2015-04-20, using tokenizations output by SyntaxNet.}
its disparity is worse.
\cite{Soni2014Factuality} used a similar parser\footnote{The older Stanford \emph{englishPCFG} model
with dependency transform (via pers.\ comm.).
}
on Twitter text;
our analysis suggests this approach may suffer from errors caused by the parser.

\ignore{
\begin{table}
    \centering
    \caption{Dependency parser accuracy (labeled recall on partial annotations)
    on 100 messages
    each from AA-aligned and white-aligned
    subcorpora. Bootstrapped standard error in parentheses.
    \label{f:parseeval}}
\end{table}
}


\section{Discussion and Conclusion}

We have presented a distantly supervised probabilistic model that employs demographic correlations of a dialect and its speaker communities to uncover dialectal language on Twitter.
Our model can also close the gap between NLP tools' performance on dialectal and standard text.

This represents a case study in dialect 
\emph{identification},
\emph{characterization},
and ultimately language technology \emph{adaptation} for the dialect.
In the case of AAE, dialect identification
is greatly assisted since AAE speakers are strongly associated with a demographic group for which highly accurate governmental records (the U.S.\ Census)
exist, which we leverage to help identify speaker communities.
The notion of non-standard dialectal language
implies that the dialect is underrepresented
or underrecognized in some way,
and thus should be inherently difficult to collect data on;
and of course, \emph{many} other language communities and groups are not necessarily officially recognized.
An interesting direction for future research would be to combine distant supervision with unsupervised linguistic models to automatically uncover such underrecognized dialectal language.

\vspace{0.1in}\noindent{\footnotesize \textbf{Acknowledgments:}
We thank Jacob Eisenstein, Taylor Jones, Anna J{\o }rgensen, Dirk Hovy, and the anonymous reviewers
for discussion and feedback.}

\bibliography{emnlp2016,brenocon}
\bibliographystyle{plainnat}  

\fullversion{
\renewcommand{\thesubsection}{\Alph{subsection}}
\section*{Appendix}
\input{appendix_body}

}

\end{document}

%% file: appendix_body.tex
\subsection{Census demographics (\S2.1)}

These four ``races'' (non-Hispanic whites, Hispanics, non-Hispanic African-Americans, and Asians)
are commonly used in sociological studies of the U.S.
The Census tracks other categories as well, such as Native Americans.  The exact options the Census uses are somewhat complicated (e.g.\ Hispanic is not a ``race'' but a separate variable); in a small minority of cases, these values do not sum to one, so we re-normalize them for analysis and discard the small fraction of cases where their sum is less than 0.5.  For simplicity, we sometimes refer
to these four variables as \emph{races}; this is a simplification since the Census considers race and ethnicity to be separate variables,
and the relationship between the actual concepts of race and ethnicity is fraught on many levels.

\subsection{Unicode ranges for emoji removal (\S4.1)}

We observed that emoji symbols often caused \emph{langid.py} to
give strange results, so we preprocessed the data (improving \emph{langid.py}'s predictions)
to remove
emoji and other symbolic-type characters by removing all characters
falling into particular Unicode ranges.  We were not able to find effective pre-existing solutions,
so developed range lists by consulting documentation on Unicode and emoji standards,
and observing message samples and how our proposed rules changed them.
See \emph{emoji.py} for our implementation.
Note that Unicode consists of 17 planes of 65,536 codepoints each; each plane contains a number of variable-sized blocks.  The first three planes are most often used today 
(Basic Multilingual, Supplemental Multilingual, Supplemental Ideographic).
We remove all characters from the following ranges.

\begin{itemize}
\item 10000--1FFFF: The entire Supplemental Multilingual Plane, which contains emoji and other symbols, plus some rarely used scripts such as Egyptian hieroglyphics.
\item 30000--10FFFF: The fourth and higher planes.
\item 02500-02BFF: A collection of symbol blocks within the Basic Multilingual Plane, including: 
     Box Drawing,
     Box Elements,
     Miscellaneous Symbols,
     Dingbats,
     Miscellaneous Mathematical Symbols-A,
     Supplemental Arrows-A,
     Braille Patterns,
     Supplemental Arrows-B,
     Miscellaneous Mathematical Symbols-B,
     Supplemental Mathematical Operators,
     Miscellaneous Symbols and Arrows.
\item 0E000--0EFFF: A ``private use'' area in the BMP, for which we observed at least one non-standard character in Twitter data (U+E056).
\item 0200B--0200D: Zero-width spaces, joiner, and nonjoiner.  The ZWJ is used to create compositional emoji.\footnote{\url{http://www.unicode.org/emoji/charts/emoji-zwj-sequences.html}}  
\item 0FE0E, 0FE0F: Variation sequences, which are invisible post-modifiers allowed for certain emoji.\footnote{\url{http://unicode.org/reports/tr51/}}
\end{itemize}

\subsection{Posterior inference via CVB0 for ensemble classifier (\S4.2.1)}

The posterior inference task is to calculate the posterior expectation of
\[ P(\theta \mid w,\phi,\alpha) \propto P(\theta\mid\alpha)
P(w\mid\theta,\phi) \]
where $\phi$ are the trained topic-word language models 
 and $\theta \sim Dir(\alpha)$ is a prior over topic proportions,
 with a fixed symmetric prior $\alpha_k=1/16$.

The $\phi$ topic-word distributions are calculated via training-time posterior inference by averaging Gibbs samples $\bar{N}_{wk} = (1/S) \sum_s $ (where $s$ indexes the last 50 samples of the Gibbs sampler),
as well as adding a pseudocount of 1 and normalizing:
\[\phi_{k,w} \propto (\bar{N}_{k,w}+1)\]
(The detailed balance theory of MCMC
implies no pseudocount should be added,
but we found it seemed to help since it prevents rare words from having
overly low posterior expected counts.)

The $\hat{\theta}$ prediction is inferred as the posterior mean given the words in the message
by using
the ``CVB0'' version of variational Bayes \citep{Asuncion2009},
which is closely related to both Gibbs sampling and EM.
It iteratively updates the soft posterior for each token position $t=1..T$,
\[ q_t(k) \propto (N_{-t,k} + \alpha_k)\ \phi_{k,w_t} \]
where $N_{-t,k} = \sum_{t' \neq t} q_{t'}(k)$ is the soft topic count from other tokens in the message.
The final posterior mean of $\theta$ is estimated as $\hat{\theta}_k = (1/T) \sum_t q_t(k)$.  
We find, similar to \cite{Asuncion2009},
that CVB0 has the advantage of simplicity and rapid convergence;
$\hat{\theta}$ converges to within absolute $0.001$ of a fixed point within five iterations on test cases.

\subsection{Syntactic dependency annotations (\S5)}

The SyntaxNet model outputs grammatical relations based on
Stanford Dependencies version 3.3.0;\footnote{Personal communication with the authors.}
thus we sought to annotate messages with this formalism,
as described in a 2013 revision to \cite{Marneffe2008DepManual}.\footnote{We only had access to the 2015 version, currently available online.  We also considered follow-up works \cite{DeMarneffe2014Universal} and \url{http://universaldependencies.org/}, deferring to \cite{Marneffe2008DepManual} when in conflict.}
For each message, we parsed it and displayed the output in the Brat
annotation software\footnote{\url{http://brat.nlplab.org/}}
alongside an unannotated copy of the message, which we added dependency edges to.
This allowed us to see the proposed analysis to improve annotation speed and conformance with the grammatical standard.  
For difficult cases, we parsed shortened, Standard English toy sentences
to confirm what relations were intended to be used to capture specific syntactic constructs.  Sometimes this clearly contradicted the annotation standards (probably due to mismatch between the annotations it was trained on versus the version of the dependencies manual we viewed); we typically deferred to the parser's interpretation in such cases.

In order to save annotation effort for this evaluation, we took a partial annotation approach: for each message, we identified the root word of the first major sentence\footnote{We take \cite{Kong2014Tweeboparser}'s view 
that a tweet consists of a sequence of one or more disconnected utterances.
We sought to exclude minor utterances like
``No'' in ``No. I do not see it'' from annotation;
in this case, we would prefer to annotate ``see.''.  A short utterance of all interjections was considered ``minor''; a noun phrase or verb-headed sentence was considered ``major.''}
in the message---typically the main verb---and annotated its immediate dependent edges.
Thus for every tweet, the gold standard included one or more labeled edges, all rooted in a single token.  As opposed to completely annotating all words in a message, this allowed us to cover a broader set of messages, increasing statistical power from the perspective of sampling from a message population.
It also alleviated the need to make fewer difficult annotation decisions - linguistic phenomena such as mistokenized fragments of emoticons, symbolic discourse markers, and (possibly multiword) hashtags.

We use the \emph{twokenize} Twitter-specific tokenizer for the messages, which separates emoticons, symbols and URLs from the text \citep{OConnor2010Tweetmotif,Owoputi2013POS}\footnote{Using Myle Ott's implementation: \url{https://github.com/myleott/ark-twokenize-py}}
and use the space-separated tokenizations as input to SyntaxNet, allowing it to tokenize further.  This substantially improves accuracy by correctly splitting contractions like ``do n't'' and ``wan na'' (following Penn and English Web Treebank conventions).  However, as expected, it fails to split apostrophe-less forms like ``dont'' and more complicated multiword tokens like ``ima'' (\emph{I am going to}, which \cite{Gimpel2011POS} sought to give a joint Pronoun-Verb grammatical category),
typically leading to misanalysis as nouns.  It also erroneously splits apart emoticons and other multi-character symbolic expressions; fortunately, these are never the the head of an utterance, so they do not need to be annotated under our partial annotation design.

We find that the Stanford Dependencies system worked well as a descriptive formalism for tweets' syntax, including AAE constructions;
for example, several cases of ``gone-V,'' ``done-V,'' and habitual \emph{be}
were analyzed as auxiliary verbs (e.g. aux(let,done) in ``I done let alot of time go by''), and the SD treatment of copulas trivially extends to null copulas.

Multiword tokens like an untokenized ``dont'' (\emph{do not})
or ``af'' (\emph{as fuck}, syntactically a PP) 
pose a challenge as well.
We adopt the following convention: 
for all incoming edges that would have gone to any of their constituent words
(had the token been translated into a multitoken sequence),
we annotate that edge as going to the token.
If there are mulitple conflicting edges---which happens if the subgraph of the constituent words has multiple roots---the earliest token gets precedence.
For example,  ``I 'm tired'' has the analysis 
nsubj(I,tired), cop('m,tired);
thus the multiword token ``Im'' in ``Im tired'' would be internally multirooted.
``I`` has priority over ``'m'',
yielding the analysis nsubj(tired,Im).

Punctuation edges (\emph{punct}) were not annotated.
We found discourse edges (\emph{discourse}) to be difficult annotation decisions,
since in many cases the dependent was debatably in a different utterance.
We tended to defer to the parser's predictions when in doubt.
The partial labeling approach does not penalize the parser if the annotator gives
too few edges, but these issues would have to be tackled
to create a full treebank in future work.

\subsection{Annotation materials}

We supply our annotations with the online materials\footnote{\url{http://slanglab.cs.umass.edu/TwitterAAE/}}
as well as working notes about difficult cases.
Annotations are formatted in \emph{Brat}'s plaintext format.